\documentclass{article}

    \PassOptionsToPackage{numbers, compress}{natbib}

 \usepackage[main, final]{neurips_2026}

\usepackage[utf8]{inputenc} 
\usepackage[T1]{fontenc}    
\usepackage{hyperref}       
\usepackage{url}            
\usepackage{booktabs}       
\usepackage{amsfonts}       
\usepackage{nicefrac}       
\usepackage{microtype}      
\usepackage{xcolor}         
\usepackage{colortbl}
\usepackage{multirow}
\usepackage{adjustbox}
\usepackage{amsmath}
\usepackage{subcaption}
\usepackage{caption}
\usepackage{graphicx}
\usepackage{wrapfig}
\usepackage{amssymb}
\usepackage{tcolorbox}
\tcbuselibrary{breakable}
\usepackage{gradient-text}   

\definecolor{highlight}{HTML}{F0F0F0}

\title{Reason in the Words You Speak: \\
Idiolectal Paraphrasing Off-Policy Traces for Reasoning Distillation in VideoLLMs}

\newcommand{\ProjectURLC}[1]{%
  {\hypersetup{hidelinks}%
   \href{#1}{{%
     \fontencoding{T1}\fontfamily{cmtt}%
     \fontseries{m}\fontshape{n}\selectfont
     \gradientRGB{#1}{150,0,24}{62,142,110}%
   }}}%
}

\author{%
  \textbf{Ji Soo Lee}$^1$ \hspace{0.2cm}
  \textbf{Jinyoung Park}$^1$ \hspace{0.2cm}
  \textbf{Seohyun Lee}$^1$ \hspace{0.2cm}
  \textbf{Jongha Kim}$^2$ \hspace{0.2cm}
  \textbf{Joonmyung Choi}$^2$ \\
  \textbf{Jinsung Yoon}$^3$ \hspace{0.2cm}
  \textbf{Hyunwoo J. Kim}$^1$\thanks{Corresponding author.} \\[0.15cm]
  $^1$KAIST, $^2$Korea University, $^3$Google Cloud AI Research \\[0.15cm]
  \texttt{\{jislee,jinyoung.park,seohyunlee,hyunwoojkim\}@kaist.ac.kr} \\
\texttt{\{jonghakim,pizard\}@korea.ac.kr} \\
\texttt{jinsungyoon@google.com}
    \\[0.65em]
  \ProjectURLC{https://mlv.kaist.ac.kr/VideoEcho-R1}%
}

\begin{document}

\maketitle
\begin{abstract}
Recent large language models achieve strong performance on complex reasoning tasks, where reinforcement learning with Group Relative Policy Optimization (GRPO) has emerged as a leading paradigm for optimizing models on self-generated trajectories.
However, the on-policy nature of GRPO bounds the model to the reasoning skills it can already produce, restricting to learn more advanced capabilities.
Prior works inject privileged reasoning traces from a stronger teacher policy to guide training, yet these traces are inherently out of distribution with respect to the student policy.
We observe that this mismatch between on-policy and off-policy causes gradient clipping on semantically critical reasoning tokens, ultimately rewarding correct answers while leaving the reasoning that justifies them unlearned.
Hence, we propose \textbf{Echo-GRPO}, a framework that lets the model reason in the words it speaks.
Rather than imitating low-probability privileged traces from the teacher model, Echo-GRPO rewrites them into the student policy's own \textit{idiolect}, that is, its own characteristic vocabulary and expression patterns, while preserving their semantics via Dual-Reference Decoding.
We instantiate this framework as \textbf{VideoEcho-R1} for video reasoning distillation, achieving consistent improvements across three multimodal LLM backbones and five benchmarks.
Finally, we show that our idiolectal paraphrasing is a plug-in module that consistently improves both RL and supervised fine-tuning frameworks for reasoning distillation, demonstrating that policy-aligned supervision extends beyond GRPO.

\end{abstract}

\section{Introduction}
Recent advances in large language models have enabled significant progress on complex reasoning tasks, where Group Relative Policy Optimization (GRPO)~\cite{shao2024deepseekmath} has emerged as an effective paradigm for reinforcing reasoning over self-generated trajectories~\cite{guo2025deepseek, team2025kimi, jaech2024openai, yu2025dapo}. 
Yet on-policy GRPO is inherently bound by its own policy formulation, from which the model can only learn from its existing reasoning skills.
To address this limitation, prior work~\cite{yan2025learning, li2025tempsamp} introduces Mixed-Policy GRPO, which injects privileged reasoning traces from a stronger teacher policy by replacing one of the sampled rollouts.

\begin{figure}[t]
    \centering
    \includegraphics[width=\linewidth,
        trim={0 1.3cm 0 0},
        clip]{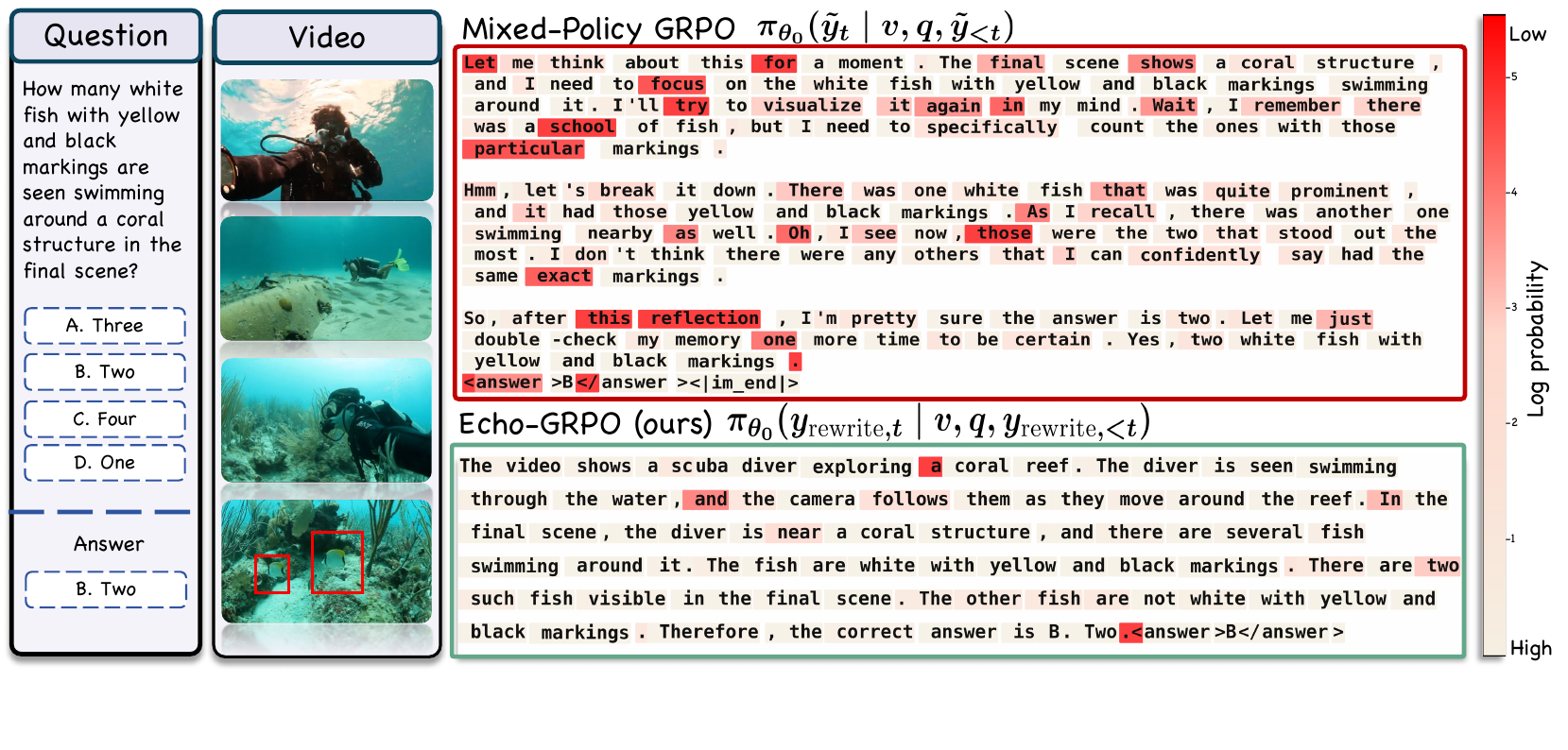}
    \caption{\textbf{Reasoning in the words you know.} Token-level log-probabilities under the policy $\pi_\theta$ (red shading indicates low probability).
    \textit{Top:} Mixed-Policy GRPO forces the policy to imitate privileged reasoning traces from a stronger teacher, leaving semantically critical ones, such as ``school of fish,'' in low-likelihood regions.
    \textit{Bottom:} Echo-GRPO paraphrases the same reasoning into the policy's native distribution (``school of fish'' $\to$ ``several fish''), preserving the semantics while keeping all tokens within the policy's distribution.}
    \label{fig:motivation}
\end{figure}

While this provides advanced guidance, it introduces a fundamental challenge.
Privileged traces are typically out-of-distribution with respect to the current policy, sampled from a stronger teacher policy rather than the student policy itself, which poses a challenge for effective learning.
This distributional gap is especially pronounced in VideoLLMs~\cite{bai2025qwen3, wang2025internvl3, seed2025seed1_5vl, ko2023large, ko2025bidirectional}, where reasoning requires complex cross-modal interactions across visuals and text, further distancing the student policy from the teacher's trace distribution.
Concretely, as shown in Fig.~\ref{fig:motivation}, most tokens in the privileged trace carry low likelihood under the policy's native distribution.
Since the student policy is unlikely to generate those tokens, the importance sampling ratio of GRPO deviates from unity, triggering trust-region clipping that suppresses their gradient updates and excludes them from training entirely.
Critically, we observe that those clipped tokens are not noise but semantically essential reasoning components, ultimately rewarding the model for the right answer while not learning the reasoning that justifies it.

To address this, we propose \textbf{Echo-GRPO}, a rewriting framework that makes the model reason in the words it speaks.
Rather than imitating out-of-distribution traces, Echo-GRPO rewrites them with respect to the student's own \textit{idiolect}, that is, its own characteristic vocabulary and expression patterns, while effectively preserving their semantics via Dual-Reference Decoding (DRD).
DRD combines two conditional distributions of the initial policy in a product-of-experts formulation: a semantic reference that constrains token selection to be faithful to the privileged trace, and a distributional reference that ensures each token remains probable under the student policy.
As illustrated in Fig.~\ref{fig:motivation}, the privileged trace describes a group of fish as a `school of fish', while the rewritten trajectory expresses the same observation as `several fish', which lies within the policy's distribution while preserving the semantic meaning.

We demonstrate that \textbf{VideoEcho-R1}, optimized with Echo-GRPO, achieves consistent improvements in reasoning distillation across three multimodal LLM backbones and five benchmarks, and our idiolectal rewriting further serves as a plug-in module that improves both reinforcement learning and supervised fine-tuning frameworks.

Our contributions can be summarized as:
\begin{itemize}
    \item We identify the underexplored failure mode of mixed-policy GRPO, where the trust-region clipping suppresses gradient updates on semantically critical reasoning tokens, rewarding correct answers while leaving the underlying reasoning unlearned.
    \item We propose \textbf{Echo-GRPO}, an idiolectal paraphrasing framework that aligns off-policy privileged traces with the policy's native distribution via Dual-Reference Decoding that promotes paraphrased trajectories to be both semantically faithful and distributionally coherent.
    \item We introduce \textbf{VideoEcho-R1}, the model trained with Echo-GRPO for video reasoning distillation, which outperforms vanilla GRPO, supervised fine-tuning, and Mixed-Policy GRPO across three multimodal LLM backbones (InternVL3.5-4B, Qwen3-VL-4B, Qwen3-VL-8B) and five video reasoning benchmarks.
    \item Idiolectal paraphrasing is a plug-in module that consistently improves reasoning distillation across both reinforcement learning frameworks and supervised fine-tuning, demonstrating that policy-aligned supervision extends well beyond vanilla GRPO.
\end{itemize}

\section{Related Works}
\noindent\textbf{Reinforcement Learning in LLM Reasoning.}
Reinforcement learning~\cite{ouyang2022training,rafailov2023direct,zhu2024self,gao2024rebel,zhou2024aligning, ko2026context, lee2025captioning, lee2025vidchain, ko2026moe} has emerged as a key driver for advancing LLM reasoning, demonstrated by recent systems such as OpenAI o1~\cite{jaech2024openai}, DeepSeek-R1~\cite{guo2025deepseek}, and Kimi-1.5~\cite{team2025kimi}.
GRPO~\cite{shao2024deepseekmath,guo2025deepseek} has emerged as a widely adopted on-policy paradigm, with subsequent work refining it to address trust-region behavior, length bias, and KL regularization~\cite{yu2025dapo, liu2025understanding, zheng2025group, park2025deepvideo}. 
A parallel line injects privileged supervision from stronger teachers to overcome the on-policy capability ceiling: Mixed-Policy GRPO~\cite{mroueh2025revisiting,yan2025learning} substitutes one rollout with an off-policy trace, LUFFY~\cite{yan2025learning} further applies regularized importance sampling to balance imitation and exploration, and OPSD~\cite{zhao2026self} minimizes per-token divergence between a privileged-context teacher and the student over student-generated rollouts.
Our work identifies a specific failure mode in this setting: gradient suppression from trust-region clipping on semantically critical tokens, and addresses it with idiolectal paraphrasing jointly faithful to the teacher's semantics and probable under the policy, directly stabilizing importance sampling ratios without requiring gradient or loss level correction.

\noindent\textbf{Reasoning in Multimodal Large Language Models.}
Recent works extend RL-based reasoning to multimodal LLMs across diverse visual tasks~\cite{huang2025vision, feng2025onethinker, li2025videochat, li2025tempsamp, wang2025timezero, chen2025exploring, wang2025time, zhang2025tinyllava, liu2026videoauto, lee2026videosearchr1}.
Vision-R1~\cite{huang2025vision} adapts R1-style RL to image reasoning, while Video-R1~\cite{feng2025video}, VideoChat-R1~\cite{li2025videochat}, and Tempsamp-R1~\cite{li2025tempsamp} extend it to video, targeting the temporal space of temporal-aware GRPO and rewards, respectively. 
Reason-RFT~\cite{tan2025reason} further explores self-reflection and staged fine-tuning for vision-language reasoning.
While these works successfully transplant RL-for-reasoning to the multimodal regime, the off-policy distillation gap is particularly acute in the video setting, where traces from large multimodal teachers lie far outside the distribution of smaller VideoLLMs, and semantic grounding in spatiotemporal evidence makes clipping-induced suppression especially damaging.

\section{Method}
In this section, we first provide a brief overview of GRPO for reasoning distillation and the assumption it relies on (Sec.~\ref{subsec:background}).
Then we identify the failure mode of Mixed-Policy GRPO, where trust-region clipping suppresses gradient updates on semantically critical reasoning tokens (Sec.~\ref{subsec:Problem}). 
To address this, we propose Echo-GRPO, an idiolectal paraphrasing framework that aligns the privileged reasoning traces into the policy's native distribution via Dual-Reference Decoding (Sec.~\ref{subsec:learn_to_reason}).

\subsection{Reasoning Distillation with GRPO}
\label{subsec:background}
\noindent\textbf{Group Relative Policy Optimization (GRPO)}~\cite{shao2024deepseekmath} serves as the underlying framework for reasoning distillation, where an off-policy reasoning trace is adopted as a supervision signal to guide optimization, also referred to as Mixed-Policy GRPO~\cite{yan2025learning}.
Given an input $(v, q)$ consisting of a video and a question, we consider two types of reasoning traces: (1) \textit{naive} reasoning traces $y \sim \pi_{\text{old}}(\cdot \mid v, q)$ sampled from the current policy, and (2) \textit{privileged} reasoning trace $\tilde{y} \sim \pi_T(\cdot \mid v, q)$ obtained from a stronger teacher policy $\pi_{T}$.
Then we combine the set of trajectories as $\mathcal{G}=\{y_i\}_{i=1}^{G - 1} \cup \{\tilde{y}\}$ where $G$ denotes the number of candidate samples.
The GRPO optimizes the following objective over $\mathcal{G}$:
\begin{equation}
\begin{split}
J_{\text{GRPO}}\left(\theta \right) = & \mathbb{E}_{v,q \sim \mathcal{D}, \{y_i\}_{i=1}^{G-1} \sim \pi_{\text{old}}\left(\cdot|v,q \right), \tilde{y} \sim \pi_T\left(\cdot|v,q \right)} \\
& \left[\frac{1}{G} \sum_{i=1}^{G} \frac{1}{|y_i|} \sum_{t=1}^{|y_i|}  \min \left( \hat{r}_{i,t}(\theta) \hat{A}_{i}, \text{clip}\left( \hat{r}_{i,t}(\theta), 1-\epsilon, 1+\epsilon \right) \hat{A}_{i} \right)\right],
\end{split}
\end{equation}
where $\epsilon$ is the clipping hyperparameter, and the importance sampling ratio is $\hat{r}_{i,t} = \frac{\pi_\theta\left(y_{i,t} | v, q, y_{i,<t} \right)}{\pi_{\text{old}}\left(y_{i, t} | v, q, y_{i, <t} \right)}$.
The advantage $\hat{A}_i$ for reasoning trace $y_i$ is computed using a group-normalized reward $\hat{A}_i = \frac{\mathcal{R}(y_i) - \mu_{\mathcal{G}}}{\sigma_{\mathcal{G}}}$, where $\mu$ and $\sigma_{\mathcal{G}}$ are the mean and standard deviation of a set of rewards from reasoning traces $\mathcal{G}$.
Following prior works~\cite{yu2025dapo,liu2025understanding}, we omit the KL divergence penalty.

The importance sampling ratio $\hat{r}_{i,t}(\theta)$ plays a central role in ensuring valid policy updates when optimizing over trajectories sampled from a previous policy close to the current policy.
Crucially, this mechanism relies on the assumption that all training trajectories in $\mathcal{G}$ lie within the behavior~(old) policy distribution, keeping importance ratios near unity and gradient updates stable.
However, adopting the privileged trace $\tilde{y}$ that is not sampled from $\pi_{\theta}$ but from a separate teacher policy $\pi_{T}$ may break this assumption.

\begin{figure*}[!t] 
    \centering
    \begin{subfigure}[h]{0.37\linewidth}
        \includegraphics[width=1.0\linewidth]{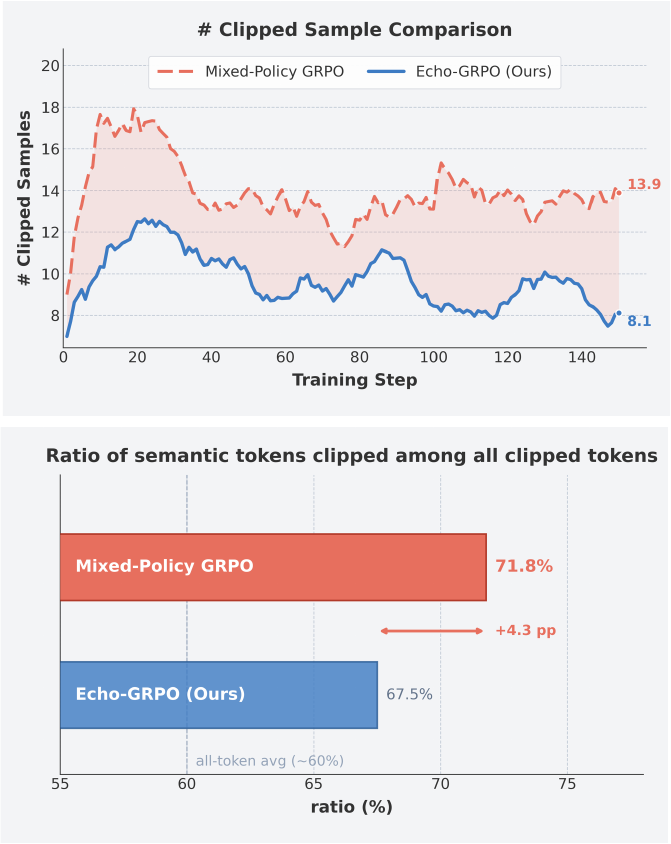}
        \caption{\textbf{Clipping rate comparison.}}
        \label{fig:ob1}
    \end{subfigure}
    \hspace{1mm}
    \begin{subfigure}[h]{0.61\linewidth}
\includegraphics[width=1.0\linewidth]{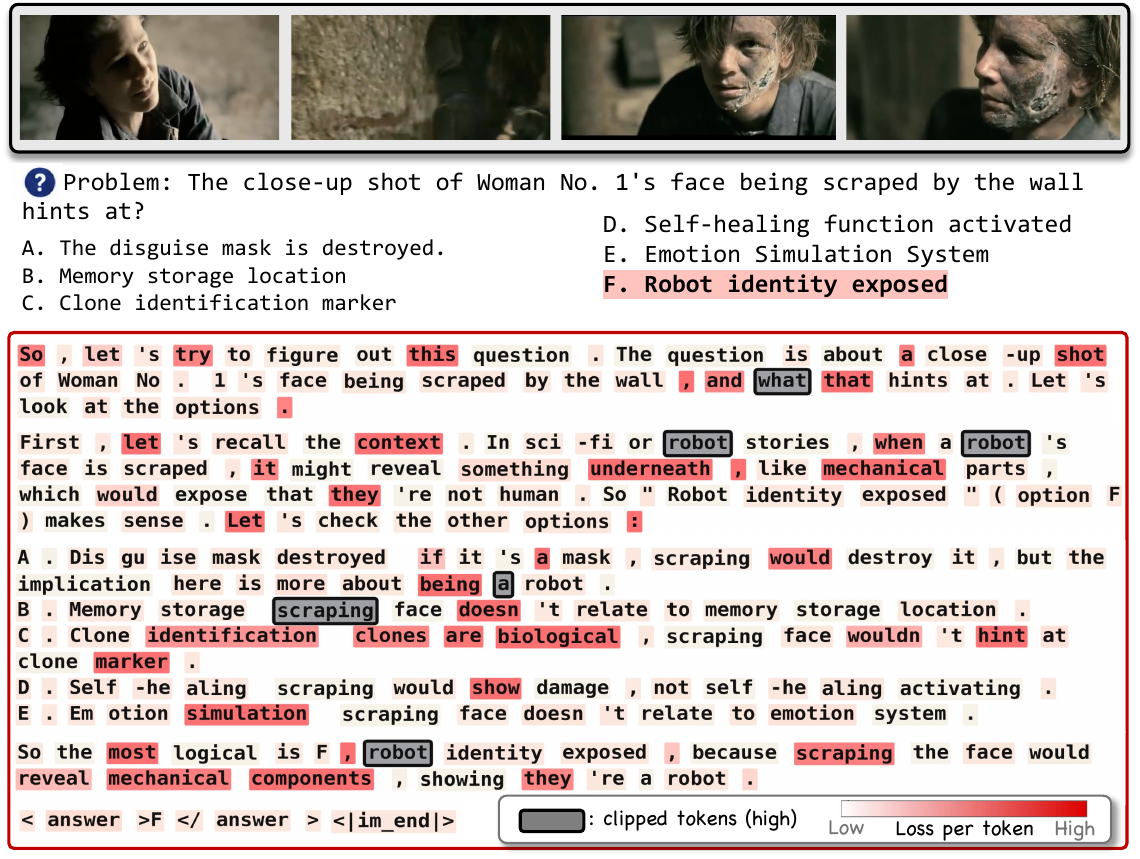}
        \caption{\textbf{Example of Semantically important token clipped.}}
        \label{fig:ob2}
    \end{subfigure}
    \caption{
    (a) (top) Number of samples containing at least one clipped token throughout training for Mixed-Policy GRPO (red), and Echo-GRPO (blue).
(bottom) Ratio of semantically important tokens (\textit{e.g.}, nouns from the question and options) among all clipped tokens for each method.
(b) Token-level visualization of clipping behavior on a representative video reasoning sample, where grey boxes indicate clipped tokens and red shading denotes per-token loss intensity.}
    \label{fig:observation}
\end{figure*}

\subsection{Clipping-Induced Suppression of Essential Reasoning}
\label{subsec:Problem}
Substituting an off-policy privileged trace $\tilde{y}$ into the GRPO objective creates a cascading failure where the distributional gap between the $\pi_{\theta}$ and $\pi_{T}$ first manifests as instability in the importance sampling ratio $\hat{r}_{i,t}(\theta)$. 
Specifically, the model is forced to update its policy towards a rigid, out-of-distribution reasoning trace, $\tilde{y}$, which it is highly unlikely to generate, \textit{i.e.}, low $\pi_{\theta}(\tilde{y}|v, q)$.

\noindent\textbf{Trust-region clipping discards a substantial fraction of samples}.
This instability propagates directly into trust-region clipping, where, when $\hat{r}_{i,t}(\theta)$ falls outside the trust region of $[1 - \epsilon, 1 + \epsilon]$, the corresponding gradient update is discarded.
In practice, since privileged traces carry positive advantages while their tokens remain low-probability under the student policy, the ratio tends to exceed $1 + \epsilon$, consistently triggering gradient suppression.
In Fig.~\ref{fig:ob1} (top), we measure the fraction of training samples containing at least one clipped token, where we find that Mixed-Policy GRPO clips a substantial fraction throughout optimization (dashed red).

\noindent{\textbf{Not all clipped tokens are noisy.}}
To better understand the impact, we visualize token-level clipping in Fig.~\ref{fig:ob2}.
Tokens excluded from gradient updates are highlighted in grey, with per-token loss intensity shaded in red.
As depicted, the entity "robot", the key referent for the answer "Robot identity exposed", is clipped from the gradient update, which indicates the trajectory received a high reward (the model arrives at the correct answer), but the gradient signal that would teach the model why the answer is correct is silently suppressed.
Similarly, as shown in Fig.~\ref{fig:ob1} (bottom), semantically important tokens (\textit{e.g.}, nouns from the question and options) account for a disproportionately large share of clipped tokens throughout training, with 71.8\% in Mixed-Policy GRPO.
\begin{wraptable}{r}{0.49\columnwidth}
    \caption{\textbf{Effect of clipping on Mixed-Policy failure.} OOD and ID refer to Out-of-distribution and In-distribution, respectively.
    Performance is averaged over three major benchmarks: VideoMMMU, MMVU, and Video-Holmes.}
    \label{tab:clip-intervention}
    \centering
    \vspace{-4pt}
    \scriptsize
    \setlength{\tabcolsep}{4pt}
    \renewcommand{\arraystretch}{1.05}
    \resizebox{\linewidth}{!}{%
    \begin{tabular}{@{}lcrr@{}}
        \toprule
        \textbf{Clipping Method}
        & \textbf{Priv. Trace}
        & \textbf{Avg.}
        & $\boldsymbol{\Delta}$ \\
        \midrule
        Mixed-Policy
        & OOD & 35.9 & -- \\
        + Unclip all tokens
        & OOD & 40.8 & +4.9 \\
        + Unclip random tokens
        & OOD & 48.4 & +12.5 \\
        + Unclip semantic tokens
        & OOD & \textbf{52.0} & \textbf{+16.1} \\
        \midrule
        \textbf{Echo-GRPO (Ours)}
        & ID & \textbf{58.2} & \textbf{+22.3} \\
        \bottomrule
    \end{tabular}%
    }
\end{wraptable}
\noindent
To directly test whether clipping semantically important tokens contributes to this failure, we vary which tokens are unclipped while keeping all other settings fixed across three major benchmarks. We compare unclipping all tokens, a randomly selected subset, and a subset of semantically important tokens. 
As shown in Tab.~\ref{tab:clip-intervention}, unclipping all tokens improves Mixed-Policy from 35.9 to 40.8, while unclipping a random subset reaches 48.4. 
Selectively unclipping semantic tokens yields the largest gain among these variants, reaching 52.0 (+16.1).
Since the random and semantic variants unclip the same number of tokens and differ only in which tokens are selected, the higher performance of semantic unclipping supports the view that clipping semantically important tokens is a key factor behind Mixed-Policy failure.

\begin{figure*}[t!]
    \centering
    \includegraphics[width=\textwidth]{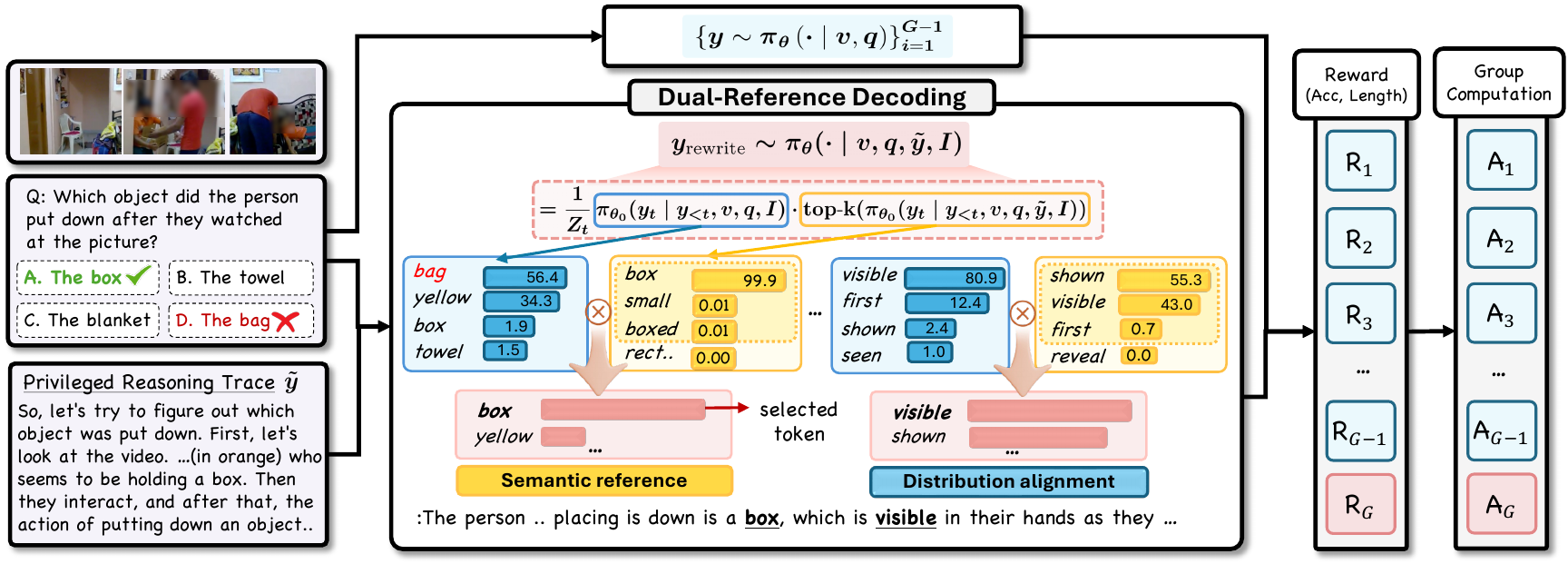}
    \caption{\textbf{Overview of Echo-GRPO.} 
    Given a video question and a privileged reasoning trace $\tilde{y}$, Echo-GRPO adopts an idiolectal paraphrased trace $y_{\text{rewrite}}$ produced by Dual-Reference Decoding (DRD).
    DRD combines two conditional distributions of $\pi_{\theta_0}$ as a product of distributions: the semantic reference (yellow) constrains token selection to be faithful to $\tilde{y}$, correcting the base policy's preference for semantically incorrect tokens (\textit{e.g.}, `bag' to `box'); while distribution alignment (blue) resolves ties among semantically equivalent candidates by preferring tokens the student policy is more likely to generate (\textit{e.g.}, `visible' over `shown'), keeping the rewritten trace aligned to its naive distribution.
    }
    \label{fig:main-fig}
\end{figure*}
\begin{figure}[t]
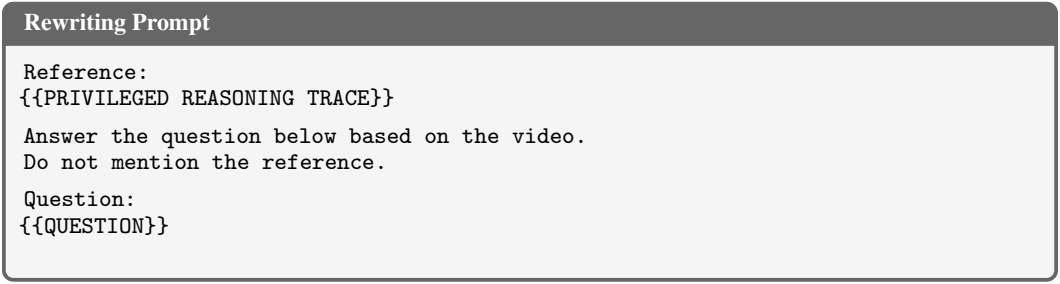

\begin{tcolorbox}[
  title=Rewriting Prompt,
  fonttitle=\bfseries\small,
  colback=gray!8,
  colframe=black!60,
  left=4pt, right=4pt, top=4pt, bottom=4pt
]
\small
\texttt{Reference:}\\
\texttt{\{\{PRIVILEGED REASONING TRACE\}\}}\\[4pt]
\texttt{Answer the question below based on the video.}\\
\texttt{Do not mention the reference.}\\[4pt]
\texttt{Question:}\\
\texttt{\{\{QUESTION\}\}}\\
\end{tcolorbox}
\caption{\textbf{Prompt for idiolectal paraphrasing.}
The privileged reasoning trace $\tilde{y}$ corresponds to (\texttt{\{\{PRIVILEGED REASONING TRACE\}\}}).
We append the question and options at the end of the prompt.
For the Question and Options we follow the format of OneThinker~\cite{feng2025onethinker}. Further details are in the supplement.}
\label{fig:prompt}
\end{figure}

\subsection{Idiolectally Paraphrasing Privileged Traces}
\label{subsec:learn_to_reason}
To address the aforementioned limitations, we propose \textbf{Echo-GRPO}, which resolves this by replacing the privileged trace $\tilde{y}$ with an idiolectally paraphrased trace $y_{\text{rewrite}}$ that preserves the semantics of $\tilde{y}$ while lying within the policy's native distribution, \textit{i.e.}, reasoning in the phrases that the model speaks with.
The goal of Echo-GRPO is to seek a rewritten trajectory $y_{\text{rewrite}}$ that satisfies two properties: (1) preserve the semantic content of $\tilde{y}$ (semantic coherence), and (2) each token should lie within the policy's native distribution (distribution alignment). 
A simple approach is sampling $y_{\text{rewrite}} \sim \pi_{\theta}(\cdot | v, q, \tilde{y}, I)$ where $I$ is a rewriting instruction prompt (shown in Fig~\ref{fig:prompt}). 
While simple rewriting conditioned on $\tilde{y}$ is effective (Sec.~\ref{ab:strategy}), it shifts the generation distribution toward the privileged trace, where we want to ensure that generated tokens lie within the policy’s native distribution \textit{without} conditioning on $\tilde{y}$.

\noindent \textbf{Dual-Reference Decoding (DRD).} To resolve this conditional shift, we introduce Dual-Reference Decoding (DRD), a product-of-experts formulation that combines two conditional distributions of $\pi_{\theta_{0}}$, where $\pi_{\theta_{0}}$ is the initial policy model. 
The first serves as a semantic reference, conditioning on the privileged trace and contributing its top-$k$ candidates, while the second acts as a distributional reference, evaluating the policy \textit{without} privileged conditioning over the full vocabulary.
At every step, the two distributions are multiplied so that the token receives high probability only when the token falls under both references.
Formally, it can be written as:
\begin{equation}
\begin{split}
\pi_{\text{rewrite}}\left(y_t \mid y_{<t}, v, q, \tilde{y}, I \right) 
& = \frac{1}{Z_t} 
\left[
\underbrace{
\pi_{\theta_0}\left(y_t \mid y_{<t}, v, q, I \right) 
}_{\text{Distributional reference}}
\cdot 
\underbrace{
\text{top-}k \left( \pi_{\theta_0}(y_t \mid y_{<t}, v, q, \tilde{y}, I) \right)
}_{\text{Semantic reference}}
\right], \\
\text{top-}k \big(\pi_{\theta_0}(y_t \mid y_{<t}, v, q, \tilde{y}, I ) \big) & = \pi_{\theta_0}(y_t \mid y_{<t}, v, q, \tilde{y}, I) \cdot \mathbf{1}_{\{y_t \in \mathcal{V}_k^{(t)}\}}, \\
\mathcal{V}_k^{(t)} &  = \underset{V \subset \mathcal{V}, |V|=k}{\arg\max} \sum_{y_t \in V} \pi_{\theta_0}(y_t \mid y_{<t}, v, q, \tilde{y}, I).
\end{split}
\end{equation}
The set $\mathcal{V}_k^{(t)}$ is the set of top-$k$ tokens under the semantic reference, and $Z_t$ is a normalization constant.
DRD actively shapes both distributions for the final probability for the current step token, of which the tokens that fall outside $\mathcal{V}_k^{(t)}$ will not be preferred as affected by the indicator $\mathbf{1}(\cdot)$.
Hence, a token is more favored when it is plausible, which is consistent with the privileged reasoning trace's semantic meaning and carries meaningful probability under the policy's native distribution.
Note that we adopt different $I$ in practice for each: the semantic reference uses the rewriting instruction prompt with the privileged trace as shown in Fig.~\ref{fig:prompt}, whereas the distributional reference uses a standard question prompt without any rewriting instruction, ensuring the output distribution reflects the model's native generation behavior.
Overall, Echo-GRPO replaces $\tilde{y}$ with $y_{\text{rewrite}}$ in a way that adheres to the on-policy assumption of GRPO, stabilizing the importance sampling ratios and ensuring that semantically critical tokens remain within the trust region, thereby enabling the model to learn meaningful reasoning components.
\section{Experiments}

\begin{table*}[!t]
    \centering
    \caption{\textbf{Comparison of baselines with general training pipelines across various benchmarks.} 
    We report performance on both Video Reasoning and Video General benchmarks using 8 frames.}
    \label{tab:main}
    \renewcommand{\arraystretch}{1.0}
    \begin{adjustbox}{width=0.97\textwidth}
    \begin{tabular}{l|c c c c c|c}
        \toprule
        \textbf{Models} 
        & \textbf{VideoMMMU} 
        & \textbf{MMVU(mc)} 
        & \textbf{Video Holmes} 
        & \textbf{VSI-Bench}
        & \textbf{VideoMME} 
        & \textbf{Avg} \\
        \midrule
        \multicolumn{7}{l}{\textit{InternVL3.5-4B}}\\
        \quad Base & 43.4 & 57.4 & 33.9 & 37.5 & 55.4 & 45.5 \\
        \quad + SFT  & 54.1 & 59.0 & 40.9 & 42.5 & 57.4 & 50.8 \\
        \quad + GRPO  & 50.4 & 61.1 & 40.0 & \textbf{46.7} & 56.5 & 50.9 \\
        \quad + SFT $\rightarrow$ GRPO  & 49.8 & 61.3 & 39.4 & 37.8 & 57.6 & 49.2 \\
        \quad + Mixed-Policy GRPO & 49.3 & 58.4 & 38.8 & 41.2 & 55.0 & 48.5 \\
        \rowcolor{highlight}
        $\textbf{VideoEcho-R1 (4B)}$ & \textbf{55.3} & \textbf{65.1} &  \textbf{41.0} & 44.0 & \textbf{57.9} & \textbf{52.7} \\
        \midrule
        \multicolumn{7}{l}{\textit{Qwen3-VL-4B}}\\
        \quad Base & 55.2 & 62.7 & 33.5 & 34.7 & 55.6 & 48.3 \\
        \quad + SFT  & 48.2 & 62.9 & 40.8 & 38.2 & 58.2 & 49.7 \\
        \quad + GRPO  & 56.8 & 68.0 & 43.3 & 50.8 & \textbf{60.7} & 55.9 \\
        \quad + SFT $\rightarrow$ GRPO  & 57.5 & 66.2 & 44.0 & 42.7 & 59.6 & 54.0 \\
        \quad + Mixed-Policy GRPO & 42.1 & 41.9 & 23.6 & 35.2 & 54.5 & 39.5 \\
        \rowcolor{highlight}
        \textbf{VideoEcho-R1 (4B)} & \textbf{59.1} & \textbf{68.8} & \textbf{46.7} & \textbf{50.9} & \textbf{60.7} & \textbf{57.2} \\
        \midrule
        \multicolumn{7}{l}{\textit{Qwen3-VL-8B}}\\
        \quad Base & 56.3 & 60.6 & 36.5 & 40.2 & 60.0 & 50.7 \\
        \quad + SFT  & 57.0 & 71.0 & 42.7 & 39.8 & 60.7 & 54.2 \\
        \quad + GRPO  & 62.6 & 71.5 & 44.7 & \textbf{51.1} & 61.0 & \textbf{58.2} \\
        \quad + SFT $\rightarrow$ GRPO  & 61.7 & 69.9 & 44.6 & 46.6 & 60.6 & 56.7 \\
        \quad + Mixed-Policy GRPO  & 60.1 & 71.5 & 43.1 & 35.8 & 57.5 & 53.6 \\
        \rowcolor{highlight}
        $\textbf{VideoEcho-R1 (8B)}$ & \textbf{63.6} & \textbf{72.3} & \textbf{45.1} & 49.1 & \textbf{61.1} & \textbf{58.2} \\
        
        \bottomrule
    \end{tabular}
    \end{adjustbox}
\end{table*}
\begin{table*}[!t]
    \centering
    \caption{\textbf{Comparison with various reasoning distillation frameworks.} 
    We report performance on both Video Reasoning and Video General benchmarks using 8 frames with Qwen3-VL-4B.
    $*$ indicates the frameworks originally proposed for LLMs, adopted to our optimization setting with minimal modifications under the same protocols for fair comparison.}
    \label{tab:distillation}
    \renewcommand{\arraystretch}{1.0}
    \begin{adjustbox}{width=0.97\textwidth}
    \begin{tabular}{l|c c c c c|c}
        \toprule
        \textbf{Method} 
        & \textbf{VideoMMMU} 
        & \textbf{MMVU(mc)} 
        & \textbf{Video Holmes} 
        & \textbf{VSI-Bench}
        & \textbf{VideoMME} 
        & \textbf{Avg} \\
        \midrule
        \quad  RL w/ SFT & 52.8 & 61.0 & 39.1 & 45.5 & 53.5 & 50.4 \\
        \rowcolor{highlight}
        \quad  \textbf{RL w/ SFT + Echo } & 57.2 & 62.1 & 40.5 & 47.1 & 58.0 & 53.0 \\
        \quad  $\text{LUFFY}^*$ & 55.2 & 62.6 & 33.3 & 50.1 & 58.3 & 51.9 \\
        \rowcolor{highlight}
        \quad  \textbf{LUFFY + Echo} & 56.0 & 63.8 & 32.8 & 50.3 & 58.3 & 52.2 \\
        \midrule
        \quad  $\text{OPSD-JSD}^*$  & 56.1 & 63.8 & 36.9 & 50.8 & 55.9 & 52.7 \\
        \quad  $\text{OPSD-IKL}^*$ & 55.9 & 63.5 & 34.1 & 49.4 & 56.0 & 51.8 \\

        \midrule
        \rowcolor{highlight}
        \quad  \textbf{Echo-GRPO} & \textbf{59.1} & \textbf{68.8} & \textbf{46.7} & \textbf{50.9} & \textbf{60.7} & \textbf{57.2} \\
        \bottomrule
    \end{tabular}
    \end{adjustbox}
\end{table*}
\begin{table*}[!t]
    \centering
    \caption{\textbf{Comparison of rewriting strategies for Echo-GRPO.}
    We compare generic paraphrasing, prompt-only student rewriting, and
    DRD variants using semantic and/or distributional references on Qwen3-VL-4B.
    Note that generic paraphrasing and prompt-only student rewriting are single-step sentence rewriting of the privilege trace with an external model and the student model, respectively.}
    \label{tab:ab-strategy}
    \renewcommand{\arraystretch}{1.0}
    \begin{adjustbox}{width=0.97\textwidth}
    \begin{tabular}{@{}lccccc|c|c@{}}
        \toprule
        \textbf{Strategy}
        & \textbf{Sem. Ref.}
        & \textbf{Dist. Ref.}
        & \textbf{V-MMMU}
        & \textbf{MMVU}
        & \textbf{VHolm}
        & \textbf{Avg.}
        & $\boldsymbol{\Delta}$ \\
        \midrule
        Mixed-Policy
        & -- & --
        & 42.1 & 41.9 & 23.6 & 35.9 & -- \\

        Generic Paraphrasing
        & -- & --
        & 52.1 & 64.5 & 41.4 & 52.7 & +16.8 \\

        Prompt-only Student Rewriting
        & -- & --
        & 58.0 & 67.8 & 44.9 & 56.9 & +21.0 \\
        \midrule
        Semantic-reference only
        & $\checkmark$ & --
        & 58.5 & 66.7 & 45.3 & 56.8 & +20.9 \\

        Distributional-reference only
        & -- & $\checkmark$
        & 57.8 & 64.0 & 42.9 & 54.9 & +19.0 \\

        \rowcolor{highlight}
        \textbf{DRD (Ours)}
        & $\checkmark$ & $\checkmark$
        & \textbf{59.1}
        & \textbf{68.8}
        & \textbf{46.7}
        & \textbf{58.2}
        & \textbf{+22.3} \\
        \bottomrule
    \end{tabular}%
    \end{adjustbox}
\end{table*}

\subsection{Experimental Settings}
We evaluate three multimodal LLM backbones: InternVL3.5-4B~\cite{wang2025internvl3}, Qwen3-VL-4B~\cite{bai2025qwen3}, and Qwen3-VL-8B, with Qwen3-VL-4B used for all ablations unless noted. We uniformly sample 8 frames per video and generate 6 rollouts per sample, replacing one with the privileged trace for distillation frameworks.
All models are optimized using the same training protocol unless otherwise specified.
Our training set consists of 2.4K samples from OneThinker-SFT-340K~\cite{feng2025onethinker}, annotated by Seed1.5-VL~\cite{seed2025seed1_5vl} as teacher policy $\pi_T$; privileged traces are used only during training and not at inference time.
We use $k{=}5$ for DRD and combine accuracy and length rewards. 
We evaluate on five benchmarks: VideoMMMU~\cite{hu2025video}, MMVU (multiple choice)~\cite{zhao2025mmvu}, Video Holmes~\cite{cheng2025video}, VSI-Bench~\cite{yang2025thinking}, and VideoMME~\cite{fu2025video}.
Full hyperparameter details are in the supplement.
\subsection{Main results}
\label{subsec:main_results}
Tab.~\ref{tab:main} reports the main results of VideoEcho-R1 across three backbones (InternVL3.5-4B and Qwen3-VL-4B and Qwen3-VL-8B), comparing against other general training pipelines on five benchmarks spanning general video understanding and reasoning. 

\noindent \textbf{VideoEcho-R1 outperforms on average across backbones and benchmarks.}
On Qwen3-VL-4B, VideoEcho-R1 achieves an average score of 57.2, surpassing vanilla GRPO by 1.3 points and SFT$\rightarrow$GRPO by 3.2 points.
On InternVL3.5-4B, VideoEcho-R1 scores 52.7 on average, outperforming GRPO by 1.8 points and SFT$\rightarrow$GRPO by 3.5 points.
On Qwen3-VL-8B, VideoEcho-R1 reaches 58.2, with consistent gains on reasoning benchmarks such as VideoMMMU (+1.0) and MMVU (+0.8) over GRPO.
On VSI-Bench, which rewards exact numerical estimations (\textit{e.g.}, object size, distance), GRPO is competitive or superior.
We conjecture that on-policy rollouts produce more consistent numerical reasoning patterns within rollout groups, which is beneficial for tasks requiring fine-grained numerical precision.
 Notably, this trade-off is partially mitigated when idiolectical rewriting is applied on top of SFT$\rightarrow$GRPO with consistent reasoning patterns (see supplement).

\noindent{\textbf{Mixed-Policy GRPO empirically degrades reasoning.}} Across all benchmarks, Mixed-Policy GRPO performs worse than vanilla GRPO with as much as a drop of 16.4 points for Qwen3-VL-4B, a 2.4 points drop for InternVL3.5-4B, and a 4.6 points drop for Qwen3-VL-8B.
This empirically confirms our observation in Sec.~\ref{subsec:Problem}, where naive injection of the off-policy privileged reasoning trace hinders effective reasoning distillation.

\begin{figure*}[t]
    \centering
    \includegraphics[width=0.94\textwidth]{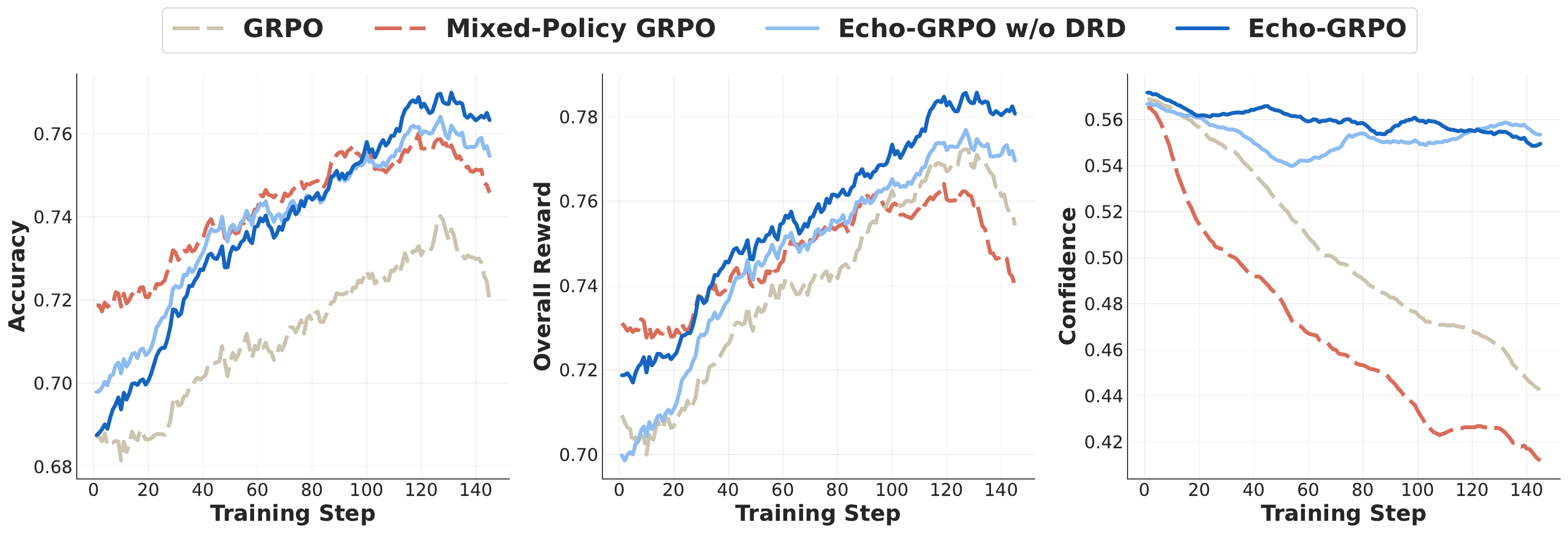}
    \caption{
        \textbf{Training Dynamics of Echo-GRPO.}
    }
    \label{fig:dynamic}
    \vspace{-0.5cm}
\end{figure*}
\subsection{Comparison with Reasoning Distillation Frameworks}
\label{subsec:distillation}
Tab.~\ref{tab:distillation} compares Echo-GRPO against three representative reasoning distillation frameworks: 
RL w/ SFT, a GRPO objective with an SFT loss on the privileged trace as ground-truth target, LUFFY, which adopts privileged traces while shaping the gradient for optimization, and the OPSD variants, which distill privileged supervision via fixed-divergence objectives. 
More details are in the supplement.

\noindent{\textbf{Echo-GRPO is the strongest among reasoning distillation methods.}}
Echo-GRPO achieves an average performance of 57.2 across five benchmarks, outperforming RL w/ SFT by 6.8 points, LUFFY by 5.3 points, OPSD-JSD by 4.5 points, and OPSD-IKL by 5.4 points.
Among the baselines, OPSD performs competitively, as it adopts divergence objectives, yet still falls short on reasoning-intensive benchmarks.
We attribute this gap to its reliance on a single privileged trace per question, which limits the diversity of reasoning patterns.
In contrast, Echo-GRPO samples various reasoning traces, exposing the model to a richer set of policy-aligned reasoning paths.

\noindent{\textbf{Idiolectal paraphrasing as plug-in module.}}
To assess whether policy-aligned supervision extends beyond Echo-GRPO, we apply our idiolectal paraphrasing to two existing frameworks. 
RL w/ SFT + Echo improves over RL w/ SFT from 50.4 to 53.0, a gain of 2.6 points.
LUFFY + Echo improves over LUFFY from 51.9 to 52.2.
The consistent gains across frameworks confirm that policy-aligned supervision with idiolectal paraphrasing is broadly beneficial and is a general plug-in applicable to existing reasoning distillation paradigms, not specific to the GRPO objective.

\subsection{Analysis}
We analyze various aspects of Echo-GRPO: (1) reduced clipping, (2) training dynamics, (3) training dynamics as a plug-in, (4) generalizability, (5) ablation on the components, (6) ablation on different rewriting strategies, (7) self-paraphrasing beyond RL, and (8) Qualitative results.
Unless specified, all analyses use Qwen3-VL-4B.

\noindent{\textbf{Reduced clipping on semantically critical tokens.}}
As shown in Fig.~\ref{fig:ob1} (top), unlike Mixed-Policy GRPO, which exhibits a substantial fraction of training samples that contain clipped tokens, suppressing gradient updates on essential reasoning components,  Echo-GRPO (blue) alleviates the clip fraction.
As visualized in Fig.~\ref{fig:ob1} (bottom), the ratio of semantically important tokens among all clipped tokens is reduced by 4.3 percentage points under Echo-GRPO (67.5\%) compared to Mixed-Policy GRPO (71.8\%). 

\noindent{\textbf{Training dynamics of Echo-GRPO.}}
Fig.~\ref{fig:dynamic} illustrates the training dynamics of overall accuracy (left), overall reward (middle), and confidence \textit{i.e.}, mean log probability per token (right),
In terms of accuracy and reward, Mixed-Policy initially achieves higher accuracy, yet shows a low learning slope, while Echo-GRPO demonstrates acceleration with sustained increase, ultimately surpassing all baselines.
Also, Echo-GRPO preserves model confidence, while GRPO exhibits a steady decline, and Mixed-Policy collapses in the early stages.

\begin{figure*}[t]
    \centering
    \includegraphics[width=0.94\textwidth]{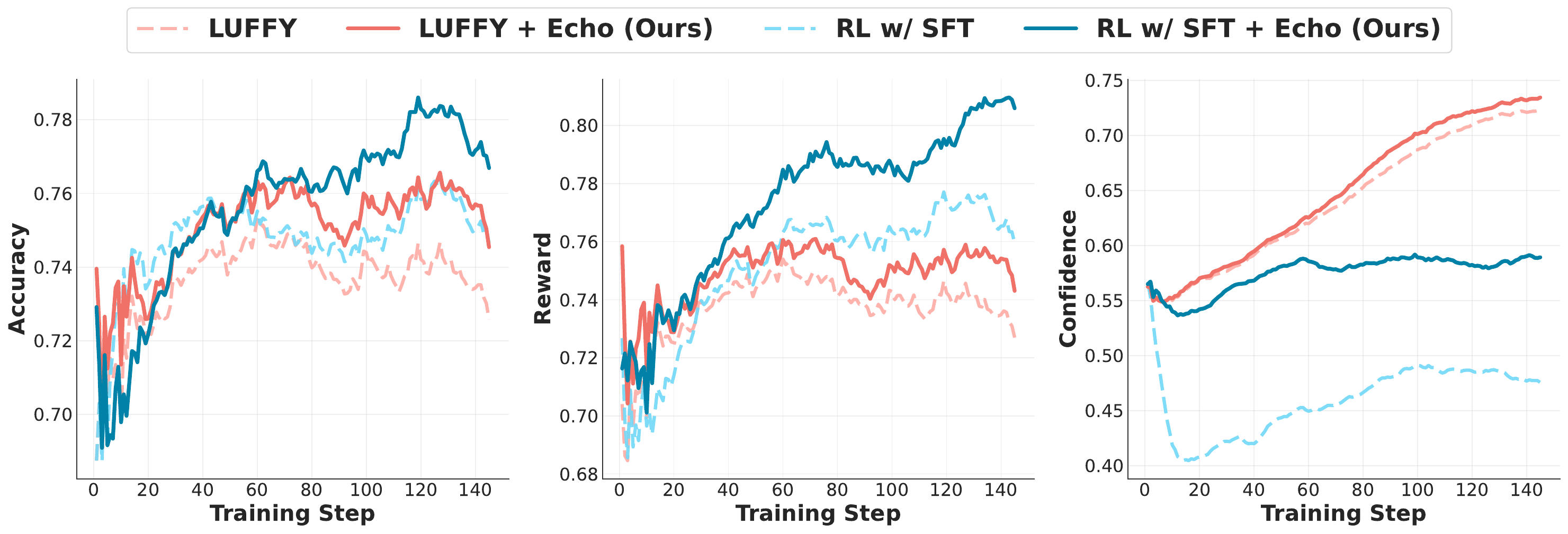}
    \caption{
        \textbf{Training Dynamics of idiolectal paraphrasing as a plug-in.}
    }
    \label{fig:plugin}
\end{figure*}

\noindent{\textbf{Training dynamics as Plug-in.}}
Fig.~\ref{fig:plugin} reveals that idiolectal paraphrasing is also effective in other reasoning distillation frameworks.
LUFFY + Echo and RL w/ SFT + Echo show consistent improvements in confidence, accuracy, and overall reward.
Both surpass their respective base frameworks throughout training, indicating idiolectal paraphrasing is broadly applicable and effective.

\begin{figure*}[!t]
    \centering
    \footnotesize
    \captionsetup{skip=3pt}
    \begin{minipage}{0.91\textwidth}
        \centering
        \begin{minipage}[t]{0.53\linewidth}
            \centering
            \includegraphics[width=\linewidth]
            {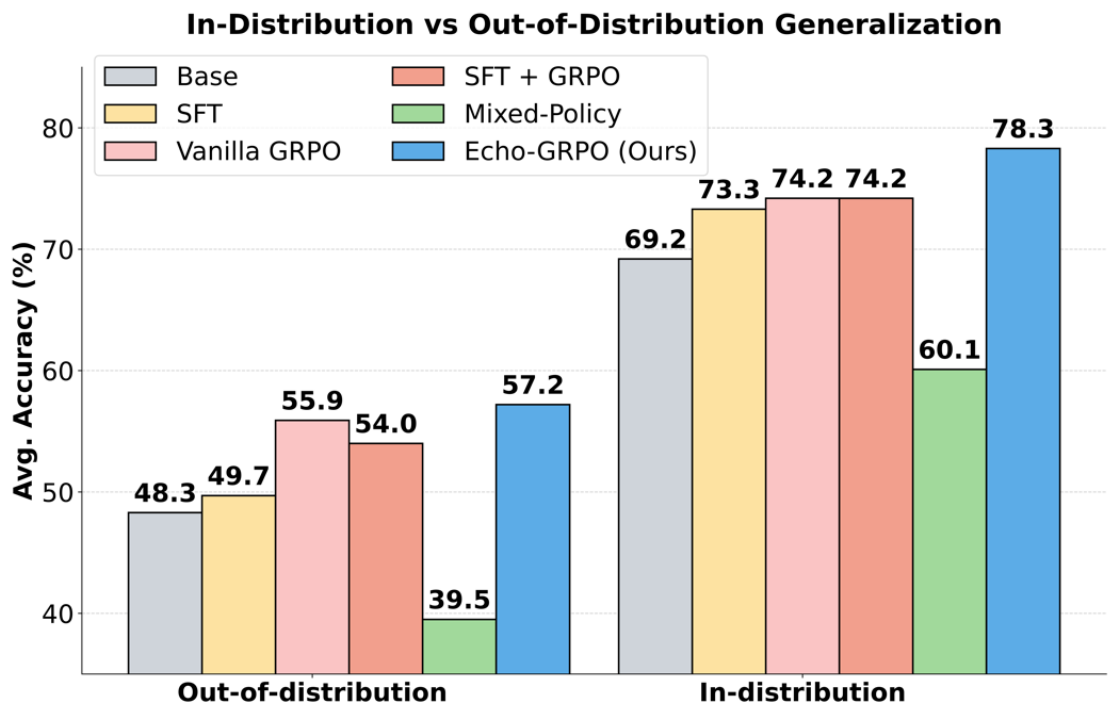}
            \captionof{figure}{\textbf{In-Distribution and Out-of-Distribution
            Generalization of Echo-GRPO}.}
            \label{fig:gen}
        \end{minipage}
        \hfill
        \begin{minipage}[t]{0.44\linewidth}
            \centering
            \includegraphics[width=\linewidth]
            {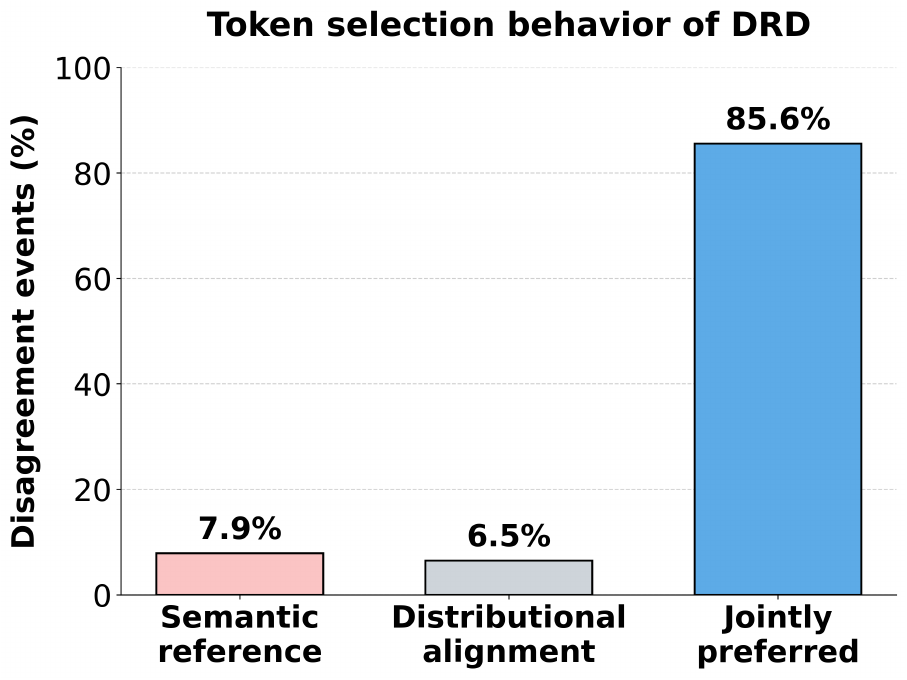}
            \captionof{figure}{\textbf{Token selection behavior of Dual-Reference Decoding}.}
            \label{fig:token-selection}
        \end{minipage}
    \end{minipage}
    \vspace{-10pt}
\end{figure*}

\noindent{\textbf{In- and out-of-distribution generalization.}}
Fig.~\ref{fig:gen} evaluates generalizability across all methods, where in-distribution refers to the held-out test set aligned with training data and out-of-distribution to general video benchmarks.
Echo-GRPO achieves best performance in both regimes (78.3 and 57.2), suggesting that our rewriting enhances reasoning rather than mere mimicking, leading to more robust generalization.

\noindent{\textbf{Ablation of rewriting strategy.}}
\label{ab:strategy}
Tab.~\ref{tab:ab-strategy} compares DRD with simpler rewriting strategies across three representative video reasoning benchmarks. 
Generic paraphrasing, where an external model (InternVL3.5-4B) rewrites the whole sentence in a single step, already improves Mixed-Policy from 35.9 to 52.7, while prompt-only \textit{student} rewriting that is single-step rewriting by the student model further reaches 56.9, showing that rewriting privileged traces toward the student's distribution is itself beneficial. 
Beyond this effect, full DRD reaches 58.2, providing an additional 1.3-point gain over prompt-only rewriting. 
Among the single-reference variants, semantic guidance performs better than distributional guidance (56.8 vs.\ 54.9), indicating the importance of preserving semantically informative tokens. 
However, combining both references in DRD consistently improves over semantic-only guidance across all three benchmarks by +0.6, +2.1, and +1.4 points, respectively, yielding the best average performance of 58.2.
This further supports jointly incorporating semantic and distributional guidance, also reflected in the training dynamics in Fig.~\ref{fig:dynamic}. Fig.~\ref{fig:token-selection} further shows that, when the two references disagree, DRD selects a token preferred by both references in 85.6\% of cases, supporting our product-of-experts formulation.

\textbf{Idiolectal paraphrasing improves SFT beyond RL.}
Tab.~\ref{tab:ab-sft} examines whether policy-aligned supervision extends to the SFT setting by replacing the privileged trace $\tilde{y}$ with the rewritten trace $y_{\text{rewrite}}$ as the SFT target.
For SFT, this improves average performance from 50.6 to 52.6.
\begin{wraptable}{r}{0.48\columnwidth}
    \centering
    \caption{\textbf{Comparison of SFT and SFT$\rightarrow$GRPO using offline
    ($\tilde{y}$) and idiolectal paraphrased traces
    ($y_{\text{rewrite}}$).}}
    \label{tab:ab-sft}
    \setlength{\tabcolsep}{4pt}
    \renewcommand{\arraystretch}{1.08}
    \resizebox{\linewidth}{!}{%
    \begin{tabular}{@{}llrrrr@{}}
        \toprule
        \textbf{Method}
        & \textbf{Priv.}
        & \textbf{V-MMMU}
        & \textbf{MMVU}
        & \textbf{VHolm}
        & \textbf{Avg.} \\
        \midrule
        \multirow{2}{*}{SFT}
        & $\tilde{y}$
        & 48.2 & 62.9 & \textbf{40.8} & 50.6 \\
        & $y_{\text{rewrite}}$
        & \textbf{56.3} & \textbf{64.0} & 37.6 & \textbf{52.6} \\
        \midrule
        \multirow{2}{*}{SFT$\rightarrow$GRPO}
        & $\tilde{y}$
        & 57.5 & 66.2 & 44.0 & 55.9 \\
        & $y_{\text{rewrite}}$
        & \textbf{58.2} & \textbf{68.8} & \textbf{49.5} & \textbf{58.8} \\
        \bottomrule
    \end{tabular}%
    }
    \vspace{-13pt}
\end{wraptable}
With SFT$\rightarrow$GRPO, performance improves with an average from 55.9 to 58.8.
This confirms that idiolectal paraphrasing is broadly effective regardless of the training objective.

\begin{figure*}[t!]
    \centering
    \includegraphics[width=0.98\textwidth]{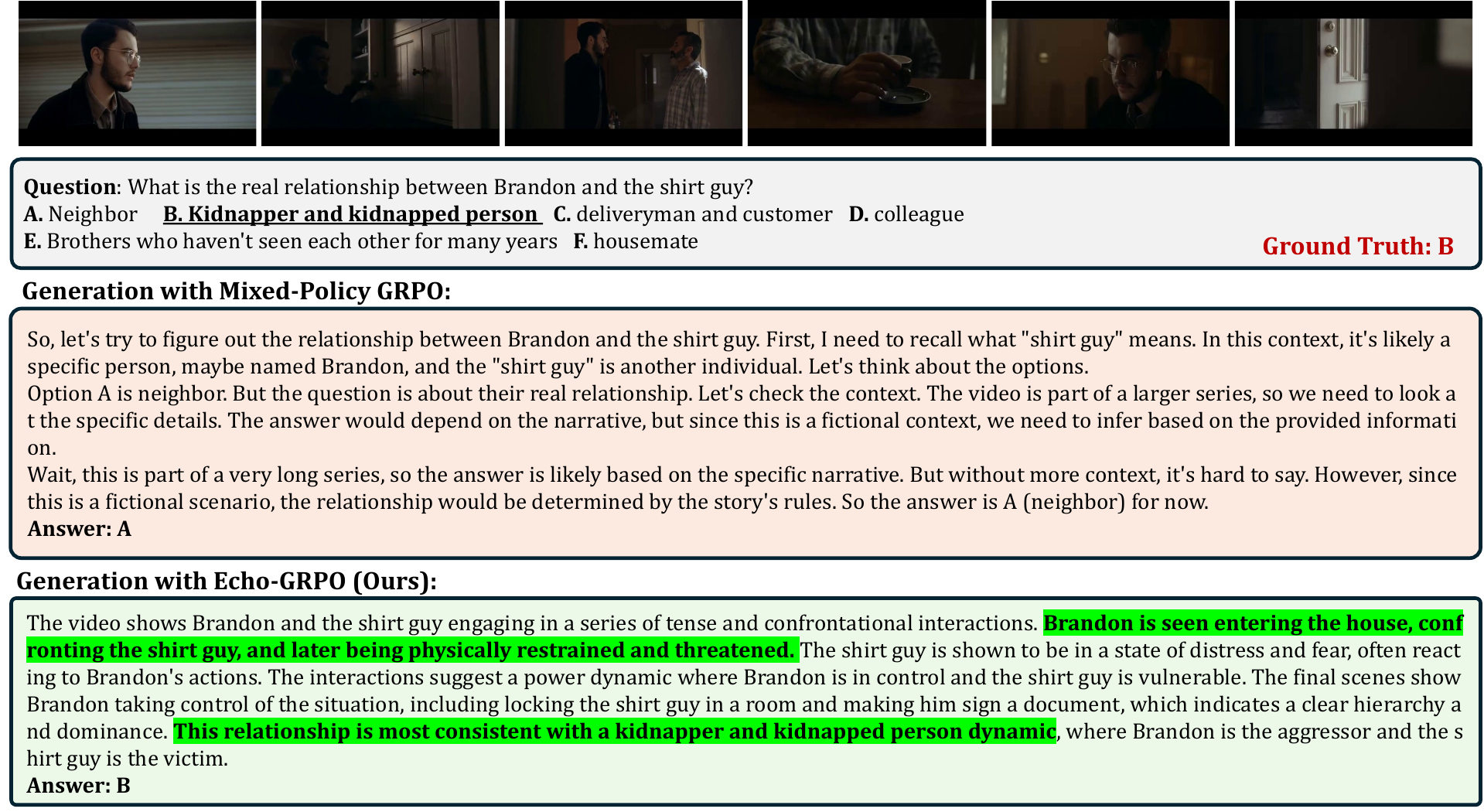}
    \caption{\textbf{Qualitative example.}   
    Mixed-Policy GRPO (top) mimics the structural format of privileged traces, defaulting to narrative speculation (``this is part of a fictional series, so the answer is likely...'') and predicting the wrong answer (A: Neighbor).
    Echo-GRPO (bottom) instead references concrete events from the video (highlighted in green) and reaches the correct answer (B: Kidnapper).} 
    \label{fig:qual}
\vspace{-0.3cm}
\end{figure*}

\noindent{\textbf{Qualitative Results.}}
Fig.~\ref{fig:qual} compares the generated reasoning trace between the Mixed-Policy GRPO and our Echo-GRPO.
Mixed-Policy GRPO fails to ground its reasoning in the video but rather tries to mimic the structural format of the privileged trace, focusing on the narrative speculation \textit{i.e.}, `this is part of a fictional series, so the answer is likely...'.
Echo-GRPO produces a visually grounded reasoning chain, identifying the concrete events (highlighted in green).
Hence, the model reaches the correct answer by reasoning with observations rather than imitating the privileged reasoning trace.

\section{Conclusion}
We identify a failure mode of mixed-policy GRPO, where trust-region clipping suppresses semantically critical tokens.
To address this, we propose Echo-GRPO, an idiolectal paraphrasing framework that rewrites privileged reasoning into the model’s native distribution via Dual-Reference Decoding.
Our VideoEcho-R1 achieves consistent improvements across multiple backbones and benchmarks, and generalizes as a plug-in module for both reinforcement learning and supervised fine-tuning.

\begin{ack}
This work was supported in part by the Institute of Information \& Communications Technology Planning \& Evaluation (IITP) grants funded by the Korea government (MSIT) Accurate and Safe Multimodal, Multilingual Personalized AI Tutors (No. RS-2024-00443251, 30\%), LG AI STAR Talent Development Program for Leading Large-Scale Generative AI Models in the Physical AI Domain (No. RS-2025-25442149, 10\%). 
This work was also supported in part by the InnoCORE program of the Ministry of Science and ICT (N10250156, 30\%) and the National Supercomputing Center with supercomputing resources including technical support (KSC-2025-CRE-0506, 30\%).
The authors declare no competing interests.
\end{ack}
{
    \small
    \bibliographystyle{unsrt}
    \bibliography{main}
}


\newpage
\appendix

\section{Experimental Details}
\label{supp:details}
We adopt three multimodal large language model backbones: InternVL3.5-4B~\cite{wang2025internvl3}, Qwen3-VL-4B~\cite{bai2025qwen3}, and Qwen3-VL-8B~\cite{bai2025qwen3}.
All models are trained with uniformly sampled 8 video frames per input.
We set the learning rate to $9\times10^{-7}$ and implement our framework in PyTorch~\cite{paszke2019pytorch}, building on the EasyR1~\cite{zheng2025easyr1} library for post-training and the LLaMAFactory~\cite{zheng2024llamafactory} framework for supervised fine-tuning.
For rollout generation and inference, we utilize vLLM~\cite{kwon2023efficient}.
For optimization, we adopt a per-step rollout batch size of 16.
Within each update, we split the rollout batch into minibatches of size 4, which defines the actor-update minibatch size.
Training is conducted on NVIDIA RTX PRO 6000 Blackwell Max-Q GPUs.

We sample 2.4k training instances from the OneThinker-SFT-340K~\cite{feng2025onethinker} dataset, focusing on video-centric and multiple-choice examples annotated by Seed1.5-VL~\cite{seed2025seed1_5vl}. 
Following our training protocol, we generate $G=6$ rollouts per sample, where one rollout is replaced with the distilled trace when distillation is applied.
We omit the KL divergence term during optimization.
For Dual-Reference Decoding, we use a top-k value of $k=5$.
All evaluations are conducted with a decoding temperature of 0.0 to ensure deterministic outputs.
In addition, we adopt LLM-based tools for the generation of the prompt and for correcting grammatical errors in the writing.

The idiolectal paraphrasing prompt is in Fig.~\ref{fig:prompt} of main. 
The instruction \texttt{Do not mention the reference} is included to prevent the model from explicitly citing or repeating the privileged trace in its response.
Without this constraint, we observe that the model tends to produce outputs that directly reference the provided trace, such as ``according to the reference'' or ``as mentioned above'', effectively copying its structure rather than genuinely paraphrasing its semantics into the policy's native distribution.
While instruction prevents 
surface-level copying of the privileged trace, it does not guarantee that generated tokens lie within the student policy's native distribution with requires our Dual-Reference Decoding.

\section{Baselines}
In this section, we describe the baseline methods used for comparison.

\noindent{\textbf{SFT}} trains the model to imitate target reasoning traces using a standard cross-entropy objective.
We adopt LoRA for parameter-efficient fine-tuning, which we found to yield strong performance compared to full-parameter training in our setting.

\noindent{\textbf{GRPO}}~\cite{guo2025deepseek} is an on-policy reinforcement learning algorithm that optimizes the model using group-wise normalized advantages computed over multiple sampled rollouts per prompt.
The policy is updated using reward-weighted likelihood ratios without relying on explicit supervision.

\noindent{\textbf{SFT $\rightarrow$ GRPO}} adopts a two-stage training strategy.
The model is first initialized via SFT (cold start) and subsequently optimized using GRPO.
For fair comparison we use 0.8K samples for SFT initialization and the remaining data for GRPO training.

\noindent{\textbf{Mixed-Policy GRPO}}~\cite{yan2025learning} incorporates off-policy supervision by injecting ground-truth reasoning traces during RL training.
Specifically, one of the sampled rollouts is replaced with a ground-truth trace, while the remaining rollouts are generated from the current policy.
This introduces a controlled off-policy signal within the GRPO framework.

\noindent{\textbf{RL w/ SFT}} jointly optimizes supervised and reinforcement learning objectives, which was introduced in \cite{yan2025learning}.
Privileged reasoning is trained with an SFT loss, while the remaining sampled rollouts are optimized using the RL objective. 
This hybrid objective enables simultaneous learning from explicit supervision and reward-driven exploration.

\noindent{\textbf{LUFFY}}~\cite{yan2025learning} is an off-policy reinforcement learning framework that leverages external reasoning traces to guide policy optimization.
It combines off-policy demonstrations with on-policy rollouts and applies regularized importance weighting to stabilize training under distribution mismatch, improving generalization beyond purely on-policy methods.

\noindent{\textbf{OPSD}}~\cite{zhao2026self} is a training paradigm in which the model learns from its own generated trajectories, adopting a single model as teacher (with privileged trace) and student by treating them as supervision signals.
It keeps the privileged trace as a distributional target in the loss function, training instead over 
student-generated rollouts via per-token KL divergence, while adopting only one trace for rollout.

\section{Evaluation Benchmarks}

We evaluate our method on a diverse suite of video question answering benchmarks that cover both general video understanding and reasoning-intensive QA scenarios.

\noindent{\textbf{Video-MMMU}}~\cite{hu2025video} evaluates the capacity of MLLMs to acquire knowledge and perform sophisticated reasoning.
Formulated around the human cognitive progression of perceiving, comprehending, and adapting, this dataset features 900 questions paired with 300 highly specialized videos.
The queries are manually crafted and distributed across six distinct fields—Medicine, Engineering, Art, Business, Science, and Humanities—requiring models to synthesize temporal visual cues with deep domain expertise.

\noindent{\textbf{MMVU (MC)}}~\cite{zhao2025mmvu} targets expert-grade, knowledge-dense video comprehension.
It encompasses 1,529 videos sourced from professional domains, accompanied by 3,000 expertly curated QA pairs.
The benchmark is broadly divided into four primary branches (Healthcare, Engineering, Science, and Humanities \& Social Sciences), which are further subdivided into 27 specific subjects.
In alignment with prior methodologies, our evaluation utilizes the multiple-choice variant, tasking models with synthesizing contextual evidence to deduce the correct candidate.

\noindent{\textbf{Video-Holmes}}~\cite{cheng2025video} is a reasoning-intensive benchmark designed to assess the complex, multi-step deductive capabilities of MLLMs.
It comprises 1,837 questions sourced from 270 expertly annotated suspense short films, with durations ranging from 1 to 5 minutes.
Diverging from traditional datasets that provide explicit context, Video-Holmes is structured around an "active seeking" paradigm.
It evaluates models across seven challenging tasks—including temporal causal inference, intention and motive chaining, and physical anomaly reasoning—requiring the model to actively locate, temporally connect, and synthesize scattered visual evidence to resolve complex causal relationships.

\noindent{\textbf{VSIBench}}~\cite{yang2025thinking} is designed to gauge visual-spatial intelligence and numerical inference.
Utilizing a collection of 288 authentic videos and over 5,000 QA pairs, it challenges models across three distinct reasoning paradigms: spatiotemporal tracking, configuration analysis, and measurement estimation.
This involves granular objectives such as estimating absolute or relative distances, counting objects, and determining appearance order, making it an ideal testbed for rigorous structural reasoning.

\noindent{\textbf{Video-MME}}~\cite{fu2025video} offers a holistic assessment of overarching video comprehension.
It contains 2,700 QA pairs tied to 900 videos that exhibit significant diversity in both content—spanning 30 sub-categories within 6 major domains—and duration.
To rigorously test temporal scalability, the benchmark categorizes videos into short (under 2 minutes), medium (4 to 15 minutes), and long (30 to 60 minutes) intervals.
To ensure a true measure of raw visual-semantic processing, we report the mean accuracy across all duration brackets without providing textual subtitles.

Overall, these benchmarks jointly assess the model’s ability to perform reasoning-driven video question answering, ranging from general understanding to structured, multi-step inference and numerical reasoning.
\section{Additional Analysis}
This section provides supplementary analyses to complement the main experimental results.
We report full benchmark results for the SFT$\rightarrow$GRPO pipeline with idiolectal paraphrased traces, and analyze the sensitivity of Dual-Reference Decoding to the top-$k$ hyperparameter.
Together, these analyses further validate the generality and robustness of Echo-GRPO across training objectives and decoding configurations.

\subsection{Full results of SFT$\rightarrow$GRPO with Idiolectal paraphrased rewriting.}
\label{supp-subsec:full-result}
\begin{table}[h]
    \centering
    \caption{\textbf{Results of SFT$\rightarrow$GRPO using offline ($\tilde{y}$) and idiolectal paraphrased traces ($y_{\text{rewrite}}$).}}
    \begin{adjustbox}{width=0.95\textwidth}
    \begin{tabular}{@{}llllllll@{}}
        \toprule
        Method & Priv. trace & V-MMMU & MMVU & VHolmes & VSI-Bench & VideoMME & Avg. \\
        \midrule
        \multirow{2}{*}{\small SFT$\to$GRPO} 
            & $\tilde{y}$ &  57.5 & 66.2 & 44.0  & 42.7 & 59.6 &  54.0 \\
            & $y_{\text{rewrite}}$ & \textbf{58.2} & \textbf{68.8} & \textbf{49.5} & \textbf{51.7} & \textbf{60.6} & \textbf{57.8} \\
        \bottomrule
    \end{tabular}
    \end{adjustbox}
    \label{supp:sft-full}
\end{table}
Tab.~\ref{supp:sft-full} reports the full benchmark results of SFT$\rightarrow$
GRPO for Qwen3-VL-4B using offline privileged traces ($\tilde{y}$) versus idiolectal paraphrased traces ($y_\text{rewrite}$).
Replacing $\tilde{y}$ with $y_\text{rewrite}$ yields consistent improvements across all five benchmarks, with an average gain of 3.8 points (54.0 $\rightarrow$ 57.8).
Notably, the gains are particularly pronounced on reasoning-intensive benchmarks such as Video Holmes (+5.5) and VideoMMMU (+0.7).
Of particular interest is VSI-Bench, where idiolectal paraphrasing achieves a substantial improvement of 9.0 points (42.7 $\rightarrow$ 51.7), comparable to Echo-GRPO (50.9), suggesting that the SFT warm-start provides a stable numerical reasoning foundation that mitigates the precision trade-off observed in vanilla Echo-GRPO.
These results confirm that the benefit of policy-aligned supervision via idiolectal paraphrasing is not specific to the GRPO objective, but extends to the SFT $\rightarrow$ GRPO pipeline as well.

\subsection{Sensitivity to top-$k$ in Dual-Reference Decoding.}
\label{supp-subsec:sensitivity}

\begin{table*}[h]
    \centering
    \caption{\textbf{Sensitivity of top-$k$ in DRD.}}
    \label{supptab:topk-ablation}
    \begin{tabular}{@{}lllll@{}}
        \toprule
        top-$k$ & VideoMMMU & MMVU & VideoHolmes & Avg. \\
        \midrule
        1 & 58.1 & 67.9 & 45.1 & 57.0 \\
        5 & \textbf{59.1} & \textbf{68.8} & \textbf{46.7} & \textbf{58.2} \\
        10 & 57.1 & 65.0 & 38.2 & 53.4 \\
        \bottomrule
    \end{tabular}
\end{table*}
A key hyperparameter of DRD is the size k of the semantic reference candidate set ${V_k}^{(t)}$.
Tab.~\ref{supptab:topk-ablation} reports performance across $k \in \{1, 5, 10\}$ on VideoMMMU, MMVU, and Video Holmes.
At $k=1$, DRD enforces strict semantic fidelity by restricting token selection to the single most probable token under the semantic reference, yet achieves a competitive average of 57.0, confirming that semantic grounding alone is beneficial.
At $k=5$, DRD achieves the best average performance of 58.2, striking the optimal balance between semantic faithfulness and distributional alignment.
At $k=10$, performance drops substantially to 53.4, suggesting that relaxing the semantic constraint too much allows the distributional reference excessive freedom, leading to semantic drift from the privileged trace and ultimately hurting reasoning quality.
These results confirm that $k=5$ is the sweet spot, where the candidate set is large enough to allow distributionally natural token choices while remaining constrained enough to preserve the semantic content of the privileged trace.

\subsection{Domain generalization of Echo-GRPO}
\begin{table}[h!]
\centering
\caption{\textbf{Effect of Echo-GRPO on text reasoning benchmarks.}}
\label{supptab:domain-generalization}
\small
\setlength{\tabcolsep}{7pt}
\renewcommand{\arraystretch}{1.1}
    \begin{tabular}{l c c c c c}
    \toprule
    \textbf{Method}
    & \textbf{AIME24}
    & \textbf{AIME25}
    & \textbf{HMMT25}
    & \textbf{Avg.}
    & $\boldsymbol{\Delta}$ \\
    \midrule
    Base
    & 16.7
    & 26.7
    & 10.0
    & 17.8
    & -- \\
    
    Mixed-Policy
    & 36.7
    & 40.0
    & 16.7
    & 31.1
    & +13.3 \\
    
    GRPO
    & 40.0
    & 40.0
    & 20.0
    & 33.3
    & +15.5 \\
    
    \textbf{Echo-GRPO}
    & \textbf{53.3}
    & \textbf{46.7}
    & \textbf{23.3}
    & \textbf{41.1}
    & \textbf{+23.3} \\
    \bottomrule
\end{tabular}
\end{table}
To evaluate whether Echo-GRPO generalizes beyond video question answering, in Tab~\ref{supptab:domain-generalization}, we conduct experiments on three widely-used for text reasoning benchmarks: AIME24~\cite{aime2024}, AIME25~\cite{aime2025}, and HMMT25~\cite{dekoninck2026matharena}. 
Following OPSD~\cite{zhao2026self}, we train on a 5K subsample of the OpenThought~\cite{guha2026openthoughts} dataset. 
We train and evaluate Qwen3-4B~\cite{qwen3} in non-thinking mode, with a maximum sequence length of 32,768 and temperature 0.0.
As shown below, Echo-GRPO achieves the best performance across all three benchmarks, improving over the base model by +23.3 points on average, over Mixed-Policy by +10.0, and over GRPO by +7.8. 
These results indicate that Echo-GRPO's policy-aligned rewriting extends beyond the video domain to text-based math reasoning, supporting the generality of the approach.

\subsection{Scale-up Experiment}
\begin{table}[h]
\centering
\caption{\textbf{Performance and clipping statistics during scaled-up training.}
The average performance covers three major benchmarks: VideoMMMU, MMVU, and Video-Holmes.
Average Clip presents the average percentage of samples clipped throughout the steps.}
\label{supptab:clip-analysis}
\small
\setlength{\tabcolsep}{4.5pt}
\renewcommand{\arraystretch}{1.1}
\resizebox{\linewidth}{!}{%
\begin{tabular}{l l c c c c c c c}
    \toprule
    \textbf{Method}
    & \textbf{Avg. Perf. ($\uparrow$)}
    & \textbf{Avg. Clip ($\downarrow$) (\%)}
    & \textbf{0--100}
    & \textbf{100--200}
    & \textbf{200--300}
    & \textbf{300--400}
    & \textbf{400--500}
    & \textbf{500+} \\
    \midrule
    Base
    & 50.5
    & --
    & --
    & --
    & --
    & --
    & --
    & -- \\
    
    Mixed-Policy
    & 51.7 {\scriptsize(+1.2)}
    & 26.1
    & 11.2
    & 9.0
    & 21.5
    & 44.6
    & 42.1
    & 28.2 \\
    
    GRPO
    & 56.4 {\scriptsize(+5.9)}
    & 7.8
    & 9.6
    & 8.4
    & 8.2
    & 6.9
    & 7.7
    & 5.7 \\
    
    \textbf{Echo-GRPO}
    & \textbf{58.2 {\scriptsize(+7.7)}}
    & \textbf{7.9}
    & \textbf{10.2}
    & \textbf{8.6}
    & \textbf{7.3}
    & \textbf{7.5}
    & \textbf{8.1}
    & \textbf{5.5} \\
    \bottomrule
\end{tabular}%
}
\end{table}
We also conducted a scaled-up experiment with approximately four times as large training set that is a total of 9k samples.
As presented in Tab~\ref{supptab:clip-analysis}, the Mixed-policy optimization shows improvement compared to the original of which it yields an average of 51.7 compared to 35.9 on the three major benchmarks of VideoMMMU, MMVU, and Video-Holmes. 
However, specifically for Mixed-policy we observe that distribution mismatch still accumulates during training: the clipping ratio rises sharply around step 200 and reaches an average of 26.1\% over the later training stages. 
In contrast, Echo-GRPO remains stable as the policy evolves, maintaining an average clipping ratio of only 7.9\%.
It also outperforms Mixed-Policy by +6.5 points on average and vanilla GRPO by +1.8 points. 
These results rule out the possibility that our previous findings were driven by the small training set. 
Even at a larger scale, Echo-GRPO provides more stable video-reasoning optimization by reducing the clipping of important tokens while keeping the rewritten trajectories aligned with the evolving student distribution.
\section{Computation Analysis of DRD}
\begin{table}[h]
  \centering
  \caption{\textbf{Performance vs.\ offline construction cost.}
  Avg.\ denotes the mean performance across VideoMMMU, MMVU, and Video-Holmes.
  Idiolectal paraphrasing is performed once during data construction;
  it introduces no additional training or inference-time overhead.}
  \begin{tabular}{@{}lccc@{}}
      \toprule
      Strategy & Avg. & $\Delta$ vs.\ GRPO
      & Offline cost$^\dagger$ \\
      \midrule
      GRPO                & 56.0 & --   & -- \\
      Echo-GRPO w/o DRD   & 56.9 & +0.9 & 9.4 ms/token \\
      Echo-GRPO w/ DRD    & \textbf{58.2} & \textbf{+2.2}
                          & 163.9 ms/token \\
      \bottomrule
  \end{tabular}
  \\[4pt]
  \small $^\dagger$One-time cost for constructing and caching the
  paraphrased training traces. Training and deployment incur no
  additional inference overhead.
  \label{supptab:cost}
\end{table}
Tab.~\ref{supptab:cost} reports the average performance across three video reasoning benchmarks and the offline construction cost of each idiolectal paraphrasing strategy.
Echo-GRPO without DRD improves over vanilla GRPO by 0.9 points with lightweight single-pass paraphrasing, while incorporating DRD yields a further 1.3-point improvement, for a total gain of 2.2 points over GRPO.
DRD requires additional computation during paraphrase generation because both semantic and distributional references are evaluated during decoding.
This computation is restricted to offline data construction: the resulting traces are generated once and cached before training.
Consequently, DRD introduces no additional decoding overhead during subsequent training or deployment.
We therefore view the construction overhead as a one-time preprocessing cost that is amortized across training, analogous to offline dataset curation in supervised fine-tuning pipelines.

\section{Broader Impact and Limitations}
\subsection{Broader Impact}
We propose Echo-GRPO, an idiolectal paraphrasing framework for reasoning distillation in videoLLMs, instantiated as VideoEcho-R1.
We believe Echo-GRPO itself does not introduce new negative societal impacts.
However, as VideoEcho-R1 is built upon pretrained multimodal language models, it may inherit biases present in pretraining data, potentially generating outputs that reflect stereotypes related to race, religion, culture, or gender.
Careful deployment aligned with responsible AI principles is necessary.

\subsection{Limitations}
Echo-GRPO introduces mild computational overhead during preprocessing due to Dual-Reference Decoding's two forward passes per step, though rewritten traces are cached prior to training and incur no inference-time cost.
Additionally, the quality of idiolectal paraphrased traces is bounded by the teacher policy's reasoning quality, and performance gains may be reduced on tasks requiring fine-grained numerical precision, as discussed in Sec.~\ref{subsec:main_results}. 
Finally, as VideoEcho-R1 is fine-tuned on top of large pretrained models, potential overlap between pretraining content and evaluation benchmarks introduces a risk of implicit data leakage.



\newpage
\end{document}